\documentclass[journal,onecolumn]{IEEEtran}
\ifCLASSINFOpdf
\else
\fi
\usepackage{epsfig}
\usepackage{graphicx}
\usepackage{amsmath}
\usepackage{amssymb}
\usepackage{commath}
\usepackage{bbding}
\usepackage{cite}
\usepackage{cleveref}
\usepackage{booktabs,multirow}

\begin{document}
%
\title{Sub-Pixel Registration of Wavelet-Encoded Images}
%
%
%

\author{Vildan~Atalay~Aydin and~Hassan~Foroosh
\thanks{Vildan Atalay Aydin and Hassan Foroosh are with the Department of Computer Science, University of Central Florida, Orlando,
FL, 32816 USA (e-mails: vatalay@knights.ucf.edu and foroosh@cs.ucf.edu).}
}

\maketitle

\begin{abstract}
Sub-pixel registration is a crucial step for applications such as super-resolution in remote sensing, motion compensation in magnetic resonance imaging, and non-destructive testing in manufacturing, to name a few. Recently, these technologies have been trending towards wavelet encoded imaging and sparse/compressive sensing. The former plays a crucial role in reducing imaging artifacts, while the latter significantly increases the acquisition speed. In view of these new emerging needs for applications of wavelet encoded imaging, we propose a sub-pixel registration method that can achieve direct wavelet domain registration from a sparse set of coefficients. We make the following contributions: (i) We devise a method of decoupling scale, rotation, and translation parameters in the Haar wavelet domain, (ii) We derive explicit mathematical expressions that define in-band sub-pixel registration in terms of wavelet coefficients, (iii) Using the derived expressions, we propose an approach to achieve in-band sub-pixel registration, avoiding back and forth transformations. (iv) Our solution remains highly accurate even when a sparse set of coefficients are used, which is due to localization of signals in a sparse set of wavelet coefficients. We demonstrate the accuracy of our method, and show that it outperforms the state-of-the-art on simulated and real data, even when the data is sparse.
\end{abstract}


\IEEEpeerreviewmaketitle

\section{Introduction} \label{sec:intro}
Image registration plays a crucial role in many areas of image and video processing, such as super-resolution  \cite{Foroosh_Chellappa_1999,Foroosh_etal_1996,shekarforoush19953d,lorette1997super,shekarforoush1998multi,berthod1994refining,shekarforoush1999conditioning,jain2008super,shekarforoush1999super}, self-localization \cite{Junejo_etal_2010,junejo2008gps,Cao_Foroosh_2007}, image annotation \cite{Tariq_etal_2017_2,Tariq_etal_2017,tariq2013exploiting,tariq2015feature,tariq2014scene,tariq2015t}, surveillance \cite{Junejo_etal_2007,Junejo_Foroosh_2008,junejo2007trajectory}, action recognition \cite{Sun_etal_2015,Ashraf_etal_2014,Ashraf_etal_2013,Shen_Foroosh_2009,shen2008view,sun2011action,ashraf2014view,shen2008action,ashraf2010view,boyraz122014action,sun2014feature,ashraf2012human}, target tracking \cite{Shu_etal_2016,shekarforoush2000multi,Milikan_etal_2017,millikan2015initialized}, shape description and object recognition \cite{Cakmakci_etal_2008,Cakmakci_etal_2008_2,ali2016character,ali2016character}, image-based rendering \cite{alnasser2006image,balci2006real,balci2006image,shen2006video}, and camera motion estimation \cite{Junejo_etal_2011,Cao_Foroosh_2007,Cao_Foroosh_2006,cao2004camera,junejo2006calibrating,cao2006self,cao2006camera,ashraf2007near}, to name a few. \\

There are various different ways that one could categorize image registration methods. In terms of functioning space, they could be either spatial domain \cite{Foroosh_2005,foroosh2001closed,foroosh2004adaptive,foroosh2003adaptive} or transform domain methods \cite{Atalay_Foroosh_2017,Balci_Foroosh_2006,Balci_Foroosh_2006_2,Foroosh_etal_2002,shekarforoush1996subpixel,foroosh2004sub,shekarforoush1995subpixel,balci2005estimating,balci2006subpixel}. On the other hand, in terms of their dependency on feature/point correspondences they may be categorized as either dependent \cite{myronenko2010point,kim2003automatic,maciel2003global,shapiro1992feature,ng2012robust,Shen09a,shen2002hammer} or independent \cite{Foroosh_2005,foroosh2001closed,foroosh2004adaptive,foroosh2003adaptive,Atalay_Foroosh_2017,Balci_Foroosh_2006,Balci_Foroosh_2006_2,Foroosh_etal_2002,shekarforoush1996subpixel,foroosh2004sub,shekarforoush1995subpixel,balci2005estimating,balci2006subpixel} of feature/point correspondences. Finally, in terms of the complexity of the image transformation, they may be categorized as linear parametric (e.g. euclidean, affine, or projective) \cite{DeCastro1987,Reddy1996,Zhang_etal_2015,Morel_2009,ZokaiOct.2005}, or semi-parametric/non-parametric diffeomorphic \cite{shen2002hammer,ashburner2007fast,klein2009evaluation,vercauteren2009diffeomorphic,thirion1998image}. The method that we propose in this paper is a parametric method that can handle full projective transformations in the Haar wavelet domain without establishing any feature or point correspondences as a preprocessing step.\\

Recently, there has been a trend in various imaging modalities and applications such as non-destructive testing and Magnetic Resonance Imaging (MRI) to adopt wavelet-encoded imaging \cite{antonini1992image,davis1999wavelet} and sparse sensing \cite{LiuNM12,Duarte_etal_2008,ma2008efficient}, with the aim of achieving better resolution, reduced distortions, higher SNR, and quick acquisition time, which are crucial for these applications. Sub-pixel registration is an integral step of various applications involving these wavelet-encoded compressive imaging technologies. Therefore, in this paper, our goal is to obtain a wavelet domain sub-pixel registration method that can achieve highly accurate results from a sparse set of wavelet coefficients. We make the following major contributions towards this goal: (i) We devise a method of decoupling scale, rotation, and translation parameters in the Haar wavelet domain, (ii) We derive explicit mathematical expressions that define in-band sub-pixel registration in terms of Haar wavelet coefficients, (iii) Using the derived expressions, we propose a multiscale approach to achieve in-band sub-pixel registration, avoiding back and forth transformations. (iv) Our solution remains highly accurate even when a sparse set of coefficients are used, due to signal energy localization in a sparse set of wavelet coefficients. Extensive experiments are used to validate our method both on simulated and real data under various scenarios.

\section{Related Work} \label{sec:related}

The earliest methods related to our work are based on image pyramids, with the aim of reducing computational time and avoiding local extrema. Examples include the work by Th\'{e}venaz \textit{et~al.} \cite{thevenaz1998pyramid}, who minimized the mean square intensity differences using a modified version of the Levenberg-Marquardt algorithm, and the work by Chen \textit{et~al.} \cite{chen_etal_2003}, who maximized the mutual information with a pyramid approach. Later, these approaches were extended to deal with local deformations in a coarse-to-fine fashion by either estimating a set of local parameters \cite{zhou2012coarse} or fitting a local model such as multi-resolution splines \cite{szeliski1994hierarchical}. Cole-Rhodes \textit{et~al.} \cite{cole2003multiresolution} proposed a method based on maximizing mutual information using stochastic gradient. Other examples of coarse-to-fine schemes are by Gong \textit{et~al.} \cite{gong2014novel}, where automatic image registration is performed by using SIFT and mutual information, and by Ibn-Elhaj \cite{ibn2009robust}, where the bispectrum is used to register noisy images. 

Hu and Acton \cite{hu2000morphological} obtain sub-pixel accuracy by using morphological pyramid structure with Levenberg-Marguardt optimization and bilinear interpolation. 
Kim et al.~~\cite{lee2012image} apply Canny edge operator in a hierarchical fashion. 
Local regions of interest of images are registered in a coarse-to-fine fashion by estimating deformation parameters by Zhou et al.~~\cite{zhou2012coarse}; 
and 
Szeliski and Coughlan \cite{szeliski1994hierarchical} present local motion field using multi-resolution splines. 

Template matching was also introduced in image registration for reducing computational cost. Ding \textit{et~al.} \cite{ding2001volume} utilized template matching with cross correlation in a spatial-domain based solution, while Rosenfeld and Vanderbrug \cite{vanderbrug1977two,rosenfeld1977coarse} used block averaging in template matching. Hirooka et al.~~\cite{hirooka1997hierarchical} optimize a small number of template points in each level of hierarchy which is selected by evaluating the correlation of images. 
In \cite{yoshimura1994fast}, Yoshimura and Kanade apply the Karhunen-Loeve expansion to a set of rotated templates to obtain eigen-images, which are used to approximate templates in the set. Tanimoto, in \cite{tanimoto1981template}, applies hierarchical template matching to reduce computation time and sensitivity to noise. Anisimov and Gorsky \cite{anisimov1993fast} work with templates which have unknown orientation, location, and nonrectangular form. 

Examples of wavelet-based methods can be summarized as follows. Turcajova and Kautsky \cite{turcajova1996hierarchical} used separable fast discrete wavelet transform with normalized local cross correlation matching based on least square fit, where spline biorthogonal and Haar wavelets outperform other types of wavelets. In \cite{kekre2012deviant},  Kekre \textit{et~al.} use several types of transforms such as discrete cosine, discrete wavelet, Haar and Walsh transforms for color image registration employing minimization of mean square error. Wang \textit{et~al.} \cite{wang2008study} improve the polynomial subdivision algorithm for wavelet-based sub-pixel image registration. 
Le Moigne \textit{et~al.} \cite{le2002automated,zavorin2005use,le1994parallel,stone1999translation,le2000use,le2000geo} have made extensive studies of various aspects of wavelet domain image registration, utilizing in particular the maxima of Daubechies wavelets for correlation based registration and multi-level optimization. 
In \cite{patil2011discrete}, Patil and Singhai use fast discrete curvelet transform with quincunx sampling for sub-pixel accuracy. Tomiya and Ageishi \cite{tomiya2003registration} minimize the mean square error; whereas, 
Wu and Chung \cite{wu2004multimodal} utilize mutual information and sum of differences with wavelet pyramids, while Wu \textit{et~al.} \cite{wu2000image} proposed a wavelet-based model of motion as a linear combination of hierarchical basis functions for image registration. 
Hong and Zhang \cite{hong2008wavelet}, combine feature-based and area-based registration, using wavelet-based features and relaxation based matching techniques, while Alam \textit{et~al.} \cite{alam2014entropy} utilize approximate coefficients of curvelets with a conditional entropy-based objective function.

These methods require transformations between spatial and transform domains, since they start at uncompressed spatial domain and use the wavelets' multiscale nature to approximate and propagate the solution from coarser to finer levels until it is refined to a good accuracy. Our method reaches the high accuracy at a coarser level with a sparse set of coefficients and no domain transformations.

\section{Sub-pixel Shifts in the Haar Domain} \label{sec:shift}
We first derive mathematical expressions that define in-band (i.e. direct wavelet-domain) shifts of an image, which will be used later for general  registration under a similarity transformation (i.e. scale, rotation, and translation) \cite{alnasser2006image}. 

\subsection{Notation}\label{sec:notation}
Table \ref{termtable} summarizes the notations used throughout the paper, to streamline the understanding of the proposed method. 
\begin{center}
\begin{table}[h] 
\centering
\caption{Notation}
\begin{tabular}{l p{0.65\linewidth}}
$I(x, y)$ & Reference image\\
$J(x,y,\sigma,\theta,t_x, t_y)$ & Sensed image to be registered to $I$\\
$\sigma$, $\theta$, $t_x$, $t_y$ & Transformation parameters to be estimated: scale, rotation angle, and shifts along the two axes, respectively\\
$A, a, b ,c$ & Wavelet transform approximation, horizontal, vertical, and diagonal detail coefficients, respectively \\
$h$ & Number of hypothetically added levels\\
$s_x$ ($s_y$)& Perceived horizontal (vertical) integer shift of wavelet coefficients after the hypothetically added levels ($h$)\\
\end{tabular} \label{termtable}
\end{table} 
\end{center}

Superscripts of $A, a, b, c$ show the level of wavelet decomposition. Subscripts $x$ and $y$ show horizontal and vertical directions, respectively; and $new$ stands for the calculated shifted coefficients. 

\subsection{In-band Shifts} \label{sec:subpixel}
We demonstrate the derived explicit mathematical expressions for an in-band translation of a given image.

Let $I(x, y)$  be a $2^N \times 2^N$ image, where $N$ is a positive integer. The Haar transform of this image consists of $N$ levels, where level $l$ holds approximation coefficient $A^l_{i,j}$ and horizontal, vertical and diagonal detail coefficients $a^l_{i,j}$, $b^l_{i,j}$, and $c^l_{i,j}$, respectively, with $l=0,...,N-1$, $i=0,...,2^l-1$ and $j=0,...,2^l-1$. 


Let, 
\begin{eqnarray}
X^l_{i,j} &=& a^{l-1}_{i,j} + b^{l-1}_{i,j} + c^{l-1}_{i,j} \nonumber \\
Y^l_{i,j} &=& -a^{l-1}_{i,j} + b^{l-1}_{i,j} - c^{l-1}_{i,j} \nonumber \\
Z^l_{i,j} &=& a^{l-1}_{i,j} - b^{l-1}_{i,j} - c^{l-1}_{i,j} \nonumber \\
W^l_{i,j} &=& -a^{l-1}_{i,j} - b^{l-1}_{i,j} + c^{l-1}_{i,j} \nonumber\\
\end{eqnarray}

Also, let $D^l_{i,j}$ be the difference between $A^0_{0,0}$ and $A^l_{i,j}$, then, $A^l_{i,j} = A^0_{0,0} + D^l_{i,j}$.

The following formula shows the relationship between $D^l_{i,j}$ and its parent level $l-1$;
\begin{equation}
D^l_{i,j} = \begin{cases}
D^{l-1}_{i/2,j/2} + X^l_{i/2,j/2}, & \text{i is even, j is even}\\
D^{l-1}_{i/2,\lfloor{j/2}\rfloor} + Y^l_{i/2,\lfloor{j/2}\rfloor}, & \text{i is even, j is odd}\\
D^{l-1}_{\lfloor{i/2}\rfloor,j/2} + Z^l_{\lfloor{i/2}\rfloor,j/2}, & \text{i is odd, j is even}\\
D^{l-1}_{\lfloor{i/2}\rfloor,\lfloor{j/2}\rfloor} + W^l_{\lfloor{i/2}\rfloor,\lfloor{j/2}\rfloor}, & \text{i is odd, j is odd}\\
0, & \text{$i = j = l = 0$}
\end{cases} 
\label{eq:D}
\end{equation}
 
Equation (\ref{eq:D}) shows that $D^l_{i,j}$, for all $l$, can be calculated only by using the detail coefficients of Haar transform iteratively, since $D^0_{0,0}=0$. We utilize $D^l_{i,j}$ to calculate the detail coefficients of the shifted image which implies that the shifting process is in-band.

We can categorize a translational shift for a 2D image into two groups for horizontal and vertical shifts where a diagonal shift can be modeled as a horizontal shift followed by a vertical one. 
Unlike the common approach of modeling sub-pixel shifts by integer shifts of some upsampled version of the given image, our method models sub-pixel shifts directly in terms of the original level coefficients. 

\textbf{Observation 3.1.} Let Haar transform of the image $I(x,y)$ have $N$ levels, with $I(x,y)$ at the $N$th level. Upsampling an image is equivalent to adding levels to the bottom of the Haar transform, and setting the detail coefficients to zero while keeping the approximation coefficients equal to the ones already in the $N$th level, $D^{N+h_0}_{i,j}=D^N_{\lfloor{i/2^{h_0}}\rfloor,\lfloor{j/2^{h_0}}\rfloor}$, where $0\leq{h_0}\leq{h}$. 

\textbf{Observation 3.2.} Shifting upsampled image by an amount of $s$ is equivalent to shifting the original image by an amount of $s/2^h$, where $h$ is the upsampling factor. 

These observations allow us to shift a reference image for a sub-pixel amount without actually upsampling it, which saves memory, reduces computation, and avoids propagating interpolation errors. 

Now, let $N'=N+h$ and $k=1+h,..., N+h$. The horizontal detail coefficients of the shifted image in case of a horizontal translation are computed from the reference image coefficients by:
\small
\begin{eqnarray} \label{eq:a} 
a^{N'-k}_{{i,j}_{new}} &=&
\sum\limits_{m=2^{k-t-1}i}^{2^{k-t-1}(i+1)-1} 
(D^N_{\lfloor{\frac{m\%{2^{N}}}{2^{h-1}}}\rfloor,\lfloor{\frac{j_1\%{2^{N}}}{2^{h-1}}}\rfloor} \nonumber\\
&+& 2\sum\limits_{n=j_1+1}^{j_2-1} 
D^N_{\lfloor{\frac{m\%{2^{N}}}{2^{h-1}}}\rfloor,\lfloor{\frac{n\%{2^{N}}}{2^{h-1}}}\rfloor} \nonumber \\
&-& 2\sum\limits_{n=j_2+1}^{j_3-1} 
D^N_{\lfloor{\frac{m\%{2^{N}}}{2^{h-1}}}\rfloor,\lfloor{\frac{n\%{2^{N}}}{2^{h-1}}}\rfloor} \nonumber\\
&-& D^N_{\lfloor{\frac{m\%{2^{N}}}{2^{h-1}}}\rfloor,\lfloor{\frac{j_3\%{2^{N}}}{2^{h-1}}}\rfloor}) \div (2 \times 4^{k-t-1})
\end{eqnarray} 
\normalsize
where,\\

\begin{eqnarray} 
j_1&=&2^{k-t-1}j+\lfloor{s_x/2^{t+1}}\rfloor \nonumber\\
j_2&=&2^{k-t-2}(2j+1)+\lfloor{s_x/2^{t+1}}\rfloor \nonumber\\
j_3&=&2^{k-t-1}(j+1)+\lfloor{s_x/2^{t+1}}\rfloor \nonumber
\end{eqnarray} 
\newline

Here, $s_x$ is the horizontal shift amount at the $(N+h)$th level (where $s$ and $h$ are calculated based on Observation $3.2$), $k$ is the reduction level, $t$ is highest power of 2 by which the shift is divisible. For the subpixel shifts, $t = 0$, since the shift amount at the hypothetically added level is always an odd integer. $t$ is essential to generalize the equation for even shifts. When $k=1$, we set the coefficients utilizing $j_2$ in Eq. (\ref{eq:a}) to $0$, since $j_2$ has a non-integer value. $a^{N'-k}_{{i,j}_{new}}$ for vertical shifts are obtained by interchanging the $a$'s with $b$'s, $i$'s with $j$'s and $m$'s with $n$'s in Eq. (\ref{eq:a}). 

By examining Eq. (\ref{eq:a}), it can be seen that each level of horizontal detail coefficients of the shifted image can be calculated using the original levels of the reference image, since $D^N_{i,j}$ is calculated in Eq. (\ref{eq:D}) using only the detail coefficients in its parent levels.

Here, we only demonstrate the formulae for horizontal detail coefficients. Approximation, vertical and diagonal detail coefficients of the shifted image can be described in a similar manner.
\section{Sub-pixel Registration} \label{sec:register}
We first demonstrate that scale, rotation, and translation can be decoupled in the wavelet domain. This is similar to decoupling of rotation and translation in Fourier domain in magnitude and phase. We then describe the proposed method to solve the decoupled registration problem for the separated parameters.

Let us assume that sensed image is translated, rotated, and scaled with respect to a reference image, in that given order. Let also $p \in I$ and $q \in J$ be two points, where $I$ and $J$ are the reference and the sensed images, respectively. The point $q$ can be defined in terms of the similarity transformation (scale, rotation, translation) and the point $p$ in terms of homogeneous coordinates as follows:
\begin{eqnarray}\label{eq:pq}
\textbf{q} &=& \textbf{S R T p}
\end{eqnarray}
\begin{eqnarray}
\begin{bmatrix}
	q_x\\
	q_y\\
	1
\end{bmatrix}
&=&
\begin{bmatrix}
\sigma && 0 && 0\\
0 && \sigma && 0\\
0 && 0 && 1
\end{bmatrix}
\begin{bmatrix}
	\cos(\theta) & -\sin(\theta) & 0 \\
	\sin(\theta) & \cos(\theta) & 0\\
	0 & 0 & 1
\end{bmatrix}
\begin{bmatrix}
1 && 0 && t_x \\
0 && 1 && t_y \\
0 && 0 && 1
\end{bmatrix}
\begin{bmatrix}
	p_x\\
	p_y\\
	1
\end{bmatrix},\nonumber
\end{eqnarray}

\normalsize

\noindent where we assume the same scale $\sigma$ for both axes. Here, $\textbf{S}$, $\textbf{R}$, and $\textbf{T}$ denote the scale, rotation, and translation matrices, and $\sigma$, $\theta$, $t_x$ and $t_y$ denote the scale factor, the rotation angle in degrees, and the translations along the two axes, respectively. 
Although we assume the order of transformations as $\textbf{S}, \textbf{R}, \textbf{T}$, we first explain rotation recovery to demonstrate the decoupling in the wavelet domain. \textbf{Algorithm 1} shows the steps of the proposed in-band registration algorithm.

\begin{table}[h]
	\textbf{Algorithm 1} \textit{In-band Registration for Similarity Transform} 
	\begin{itemize}  \itemsep1pt  \parskip0pt \parsep0pt
		\item[$\diamond$] \textit{Input}: $I(x,y)$, $J(x,y,\sigma, \theta, t_x, t_y)$
		\item[$\diamond$] \textit{Objective}: Find similarity transform parameters
		\item[$\diamond$] \textit{Output}: $translation, rotation, scale$
		\item[$\blacktriangleright$] Generate wavelet coefficients of both images
		\item[$\blacktriangleright$] \textit{\textbf{Scale recovery}} using curvature radius on coefficients
		\begin{itemize}
			\item Rescale sensed detail coefficients to the size of reference coefficients
		\end{itemize}
		\item[$\blacktriangleright$] \textit{\textbf{Rotation recovery}}  using angle histograms of coefficients
	    \begin{itemize}
	    	\item Rotate sensed detail coefficients for $- estimate$
	    \end{itemize}
		\item[$\blacktriangleright$] \textit{\textbf{Translation recovery}}  using in-band wavelet coefficient relationship (Section \ref{sec:subpixel})
	\end{itemize}
\end{table}

\subsection{Rotation Recovery}\label{sec:rotatoin}
Let $\textbf{a}$ and $\textbf{b}$ denote wavelet coefficients of the input images as in Section \ref{sec:shift}, where subscripts $I$ and $J$ stand for the images. Wavelet transform of Eq. (\ref{eq:pq}), can be defined as follows:
\begin{eqnarray}\label{eq:rotwave}
\mathbf{a}_J = \sigma \cos(\theta) \mathbf{a}_I - \sigma \sin(\theta) \mathbf{b}_I \\ \nonumber
\mathbf{b}_J = \sigma \sin(\theta) \mathbf{a}_I + \sigma \cos(\theta) \mathbf{b}_I
\end{eqnarray}

Eq. (\ref{eq:rotwave}) shows the relationship between the Haar wavelet coefficients of two images under similarity transformation, and indicate that the rotation and scale can be separated from translation, since translation parameters do not appear in these equations. In order to recover the rotation and scale independently, we also need to decouple $\sigma$ and $\theta$. One can see from Eq. (\ref{eq:rotwave}) that dividing $\textbf{b}_J$ by $\textbf{a}_J$ eliminates the scale term, the result of which is an approximation to the slopes of local image gradients using Haar coefficients, since Haar coefficients can be viewed as an estimate of partial derivatives. To obtain an initial estimate of the rotation angle, we use wavelet thresholding \cite{Donoho1994} before finding the local slopes. This will both reduce noise and sparsify the coefficients. We then find an initial estimate of the rotation angle $\theta$ by maximizing the following cross-correlation:
\begin{equation} \label{eq:theta}
\hat \theta = \operatorname*{arg\,max}_\theta (h_I \star h_J(\theta))
\end{equation}
\noindent where $\star$ denotes the cross-correlation, and $h_I$ and $h_J$ are the histogram of wavelet-coefficient slopes (HWS) for the thresholded coefficients, which we define as follows:
\begin{equation}
h_{\tt img} = \sum_{i=1}^{k} \arctan (\dfrac{\mathbf{b}_{\tt img}(i)}{\mathbf{a}_{\tt img}(i)})
\end{equation}
\noindent where $k$ is the number of bins , and the subscript ${\tt img}\in\{I, J\}$. We then refine the initial estimate $\hat \theta$, in the range $\hat \theta \pm 5^\circ$ to get the best estimate $\hat \theta^*$:
\begin{equation}
\hat \theta^* = \operatorname*{arg\,min}_{\hat\theta} || \mathbf{a}_J - \mathbf{R}\mathbf{a}_I ||_{2} + || \mathbf{b}_J - \mathbf{R}\mathbf{b}_I ||_{2}
\end{equation}

\subsection{Scale Recovery}\label{sec:scale}
Since we already demonstrated that scale, rotation and translation can be decoupled in wavelet domain, we can perform scale estimation independently of rotation and translation. Let us assume that the two images have a scale ratio of $\sigma$.
%
%
Then, the mean curvature radius calculated on thresholded wavelet coefficients would provide an accurate estimate of the scale factor:
\begin{eqnarray}
\hat \sigma \!=\!  \frac{1}{2}\!\left(\dfrac{\dfrac{1}{M_I}\sum_{i=1}^{M_I}{{\cal R}(\mathbf{a}_I(i))}}{\dfrac{1}{M_J}\sum_{i=1}^{M_J}{{\cal R}(\mathbf{a}_J(i))}}+\dfrac{\dfrac{1}{M_I}\sum_{i=1}^{M_I}{{\cal R}(\mathbf{b}_I(i))}}{\dfrac{1}{M_J}\sum_{i=1}^{M_J}{{\cal R}(\mathbf{b}_J(i))}}\right)
\end{eqnarray}
\noindent where ${\cal R}$ shows the radius of curvature.

\subsection{Translation Recovery}\label{sec:translation}
Once the scale and rotation parameters are recovered and compensated for, the translations $t_x$ and $t_y$ along the two axes can be recovered independently by maximizing the following normalized cross-correlation function:
\begin{eqnarray}\label{eq:xcorr}
&&\{{\hat t_x},{\hat t_y}\}=\operatorname*{arg\,max}_{t_x,t_y} \nonumber \\ 
&&\frac{\sum_{x,y}{(\mathbf{a}_I(x+t_x,y+t_y) \mathbf{a}_J(x',y'))}}{\sqrt{\sum_{x,y}{(\mathbf{a}_I(x+t_x,y+t_y))^2}}\sqrt{\sum_{x,y}{(\mathbf{a}_J(x',y'))^2}}}+ \nonumber\\
&&\frac{\sum_{x,y}{(\mathbf{b}_I(x+t_x,y+t_y) \mathbf{b}_J(x',y'))}}{\sqrt{\sum_{x,y}{(\mathbf{b}_I(x+t_x,y+t_y))^2}}\sqrt{\sum_{x,y}{(\mathbf{b}_J(x',y'))^2}}}
\end{eqnarray}
%
where $\mathbf{a}_I(x+t_x,y+t_y)$ and $\mathbf{b}_I(x+t_x,y+t_y)$ are the shifted versions of the reference detail coefficients (corresponding to $\mathbf{a}_{new}$ or $\mathbf{b}_{new}$ in the derivations of Section \ref{sec:subpixel}), calculated using Eq. (\ref{eq:a}) (or the equivalent for the vertical coefficients); and $\mathbf{a}_J(x',y')$ and $\mathbf{b}_J(x',y')$ are the sensed image detail coefficients after rotation and scale compensation. 

Observation 3.2 implies that sub-pixel registration for wavelet-encoded images can be performed directly in the wavelet domain without requiring inverse transformation. Furthermore, if the encoded image is also compressed (e.g. only a sparse set of detail coefficients are available), one can still perform the registration. The latter could be for instance a case of compressed sensing imager based on Haar wavelet sampling basis. To maximize the cost function in Eq. (\ref{eq:xcorr}), we use a branch and bound (BnB) algorithm, where split of rectangle areas in BnB are decided based on the two maximum cross correlations of four bounds \cite{Marti2011}. 

\begin{table}[h]
	\textbf{Algorithm 2} \textit{Sub-pixel Shifts Estimation} 
	\begin{itemize}  \itemsep1pt  \parskip0pt \parsep0pt
		\item[$\diamond$] \textit{Input}: $I(x,y)$, $J(x', y', t_x, t_y)$  (scale and rotation corrected images)
		\item[$\diamond$] \textit{Objective}: Find translational registration parameters
		\item[$\diamond$] \textit{Output}: $(t_x, t_y)$
		\item[$\blacktriangleright$] Initialize bounds for sub-pixel shift estimates to $[-1, +1] \times [-1, +1]$.
		\item[$\blacktriangleright$] Do:
		\begin{itemize}
			\item[$\circ$] Generate horizontal/vertical detail coefficients of shifted versions of the reference image $\textbf{a}_I(x+t_{x_i},y+t_{y_i})$ and $\textbf{b}_I(x+t_{x_i},y+t_{y_i})$, using Eq. (\ref{eq:a}) (similar equation for vertical), where $t_{x_i}$ and $t_{y_i}$ are the bounds at iteration $i$.
			\item[$\circ$] Update bounds (reduce rectangles to half in size) based on the peak of cross correlation in Eq. (\ref{eq:xcorr}) for detail coefficients of shifted reference image and sensed image.
		\end{itemize}
		until maximum cross-correlation in Eq. (\ref{eq:xcorr}) exceeds $\tau$ for an estimated bound, where $\tau$ is an accuracy measure (tolerance) for cross correlation.
	\end{itemize}
\end{table}

\textbf{Algorithm 2} demonstrates the main steps of the proposed method for translation recovery. Shifted horizontal/vertical detail coefficients for the updated bounds are calculated for a specified level ($k$) using Eq. (\ref{eq:a}) (similar equation for the vertical coefficients), followed by application of maximization of Eq. (\ref{eq:xcorr}).

When the algorithm converges within an $\epsilon$ distance to the true solution, it often starts osculating. So, as a modification to a general branch and bound method, we take the mid-point of osculations as the solution, which often happens to be the true solution.

The method requires the knowledge of $A^0_{0,0}$ for in-band shifts which may limit the approach to image sizes of $2^N\times2^N$. However, the solution can be generalized to images with arbitrary sizes by simply applying the method to a subregion of size $2^N\times2^N$ of the original images.

\begin{center}
	\begin{figure}[t]
		\centering
		\begin{minipage}{.3\linewidth}
			\centering
			\includegraphics[width=4.1cm]{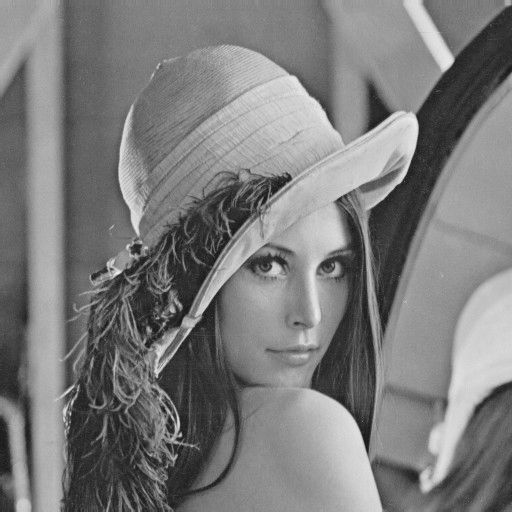}
			\centerline{\textbf{a}}
		\end{minipage}
		\hfill
		\begin{minipage}{.3\linewidth}
			\centering
			\includegraphics[width=4.1cm]{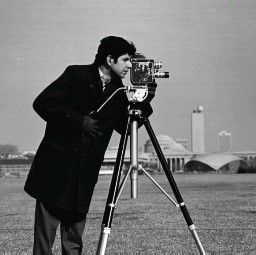}
			\centerline{\textbf{b}} 
		\end{minipage}
		\hfill
		\begin{minipage}{.3\linewidth}
			\centering
			\includegraphics[width=4.1cm]{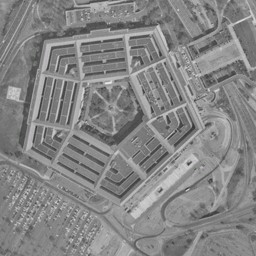}
			\centerline{\textbf{c}} 
		\end{minipage}
		\\
		\begin{minipage}{.3\linewidth}
			\centering
			\centerline{\includegraphics[width=4.1cm]{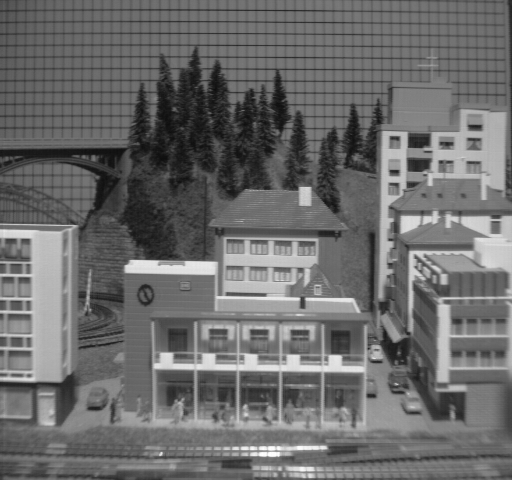}}
			\centerline{\textbf{d}} 
		\end{minipage} 
		\hfill
		\begin{minipage}{.3\linewidth}
			\centering
			\centerline{\includegraphics[width=4.1cm]{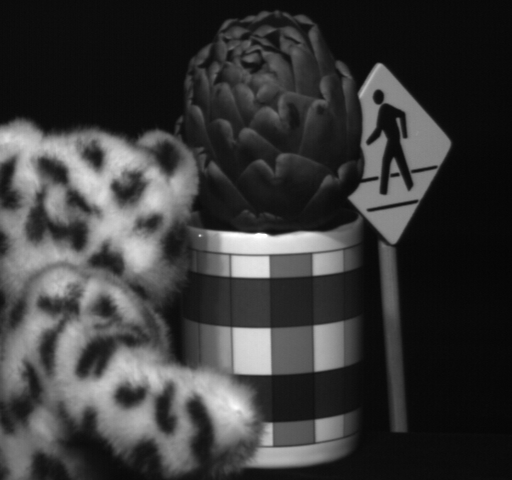}}
			\centerline{\textbf{e}} 
		\end{minipage} 
		\hfill
		\begin{minipage}{.3\linewidth}
			\centering
			\centerline{\includegraphics[width=4.9cm]{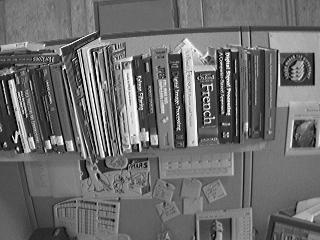}}
			\centerline{\textbf{f}} 
		\end{minipage} 
		\caption{Examples of simulated and real world images used for experiments. {\textbf{a}} {Lena} {\textbf{b}} {Cameraman} {\textbf{c}} {Pentagon} {\textbf{d}} {CIL - horizL0 \cite{CMU}} {\textbf{e}} {Artichoke - 1 \cite{CMU}} {\textbf{f}} {MDSP - Bookcase 1 \cite{UCSC}}}	\label{fig:testim}	
	\end{figure} 
\end{center}
\section{Experimental Results} \label{sec:results}
To demonstrate the accuracy of our algorithm, we performed extensive experiments on both simulated and real data. In order to simulate reference and sensed images, a given high resolution image is shifted (using bicubic interpolation) and rotated, then both images are downsampled, which is a common technique employed in state-of-the-art literature \cite{Foroosh_etal_2002}, \cite{vandewalle2007super}. If different scale are assumed, then the sensed image is also scaled further. We performed thorough comparisons with state-of-the-art methods, which were given the same input images, and results were evaluated by measuring alignment errors. Fig. \ref{fig:testim} shows some of the standard test images together with the real data obtained from \cite{CMU} and \cite{UCSC}. Captions for real data indicate the dataset and specific image names utilized as reference image. 

\subsection{Validation on Simulated Data}\label{sec:simdata}
Here, we first performed experiments on translation, rotation and scale recovery separately. We then carried out tests for combination of transformations. 

\begin{center}
	\begin{table*}[!ht] \setlength{\tabcolsep}{2pt}
		\centering
		\caption{Comparison of the proposed method with other baseline methods in estimated shifts, PSNR, and MSE.}\label{tab:GT}
		\scriptsize
		\begin{tabular}{l@{\hspace{7pt}} *{14}{r}}
			\toprule
			\multirow{2}{*}{Image}& \multirow{2}{*}{Exact shift} & \multicolumn{3}{c}{Keren \cite{keren1988image}} & \multicolumn{3}{c}{Guizar \cite{Guizar-Sicairos_2008}} & \multicolumn{3}{c}{Szeliski \cite{szeliski1994hierarchical}} & \multicolumn{3}{c}{Proposed} \\
			\cline{3-14}\\[-2ex]
			& & Est. & PSNR & MSE & Est. & PSNR & MSE & Est. & PSNR & MSE & Est. & PSNR & MSE \\
			\midrule
			\multicolumn{1}{l}{\multirow{2}[8]{*}{\textit{a}}} & 0.5 0.5 & 0.4878 0.5427 & 56.91 & 0.11 & 0.56 0.53 & 50.05 & 0.56 & 0.5017 0.5009 & 80.91 & 0 & 0.5 0.5 & Inf & 0 \\
			\multicolumn{1}{l}{} & 0.25 -0.125 & 0.2456 -0.1212 & 72.23 & 0.003 & 0.29 -0.16 & 53.10 & 0.28 & 0.2518 -0.1243 & 80.67 & 0 & 0.25 -0.125 & Inf & 0 \\
			\multicolumn{1}{l}{} & -0.375 -0.4 & -0.3826 -0.4146 & 63.99 & 0.02 & -0.42 -0.42 & 52.67 & 0.31 & -0.3732 -0.3990 & 80.36 & 0 & -0.375 -0.4023 & 82.59 & 0 \\
			\multicolumn{1}{l}{} & -0.625 0.75 & -0.6958 -0.8268 & 47.81 & 0.95 & -0.70 0.81 & 48.00 & 0.90 & -0.6231 0.7508 & 80.17 & 0 & -0.625 0.75 & Inf & 0 \\
			\hline \\ [-2ex]
			\multicolumn{1}{l}{\multirow{2}[8]{*}{\textit{b}}} & 0.33 -0.33 & 0.3347 -0.3008 & 54.19 & 0.24 & 0.27 -0.33 & 45.91 & 1.61 & 0.3275 -0.3316 & 72.49 & 0.003 & 0.3281 -0.3438 & 60.74 & 0.05 \\
			\multicolumn{1}{l}{} & 0.167 0.5 & 0.1641 0.6154 & 42.08 & 3.91 & 0.11 0.55 & 44.78 & 2.10 & 0.1633 0.4977 & 69.04 & 0.007 & 0.1719 0.5 & 67.76 & 0.01 \\
			\multicolumn{1}{l}{} & -0.875 -0.33 & -0.8639 -0.2986 & 52.58 & 0.35 & -0.92 -0.33 & 48.44 & 0.91 & -0.8783 -0.3316 & 70.51 & 0.005 & -0.875 -0.3438 & 60.40 & 0.06 \\
			\multicolumn{1}{l}{} & -0.125 0.67 & -0.1309 0.8230 & 39.53 & 7.01 & -0.08 0.75 & 42.94 & 3.20 & -0.1277 0.6695 & 72.62 & 0.003 & -0.125 0.6719 & 77.60 & 0.001 \\
			\bottomrule
		\end{tabular}%
	\end{table*}%
\end{center}


\begin{center}
	\begin{table}[!h]\setlength{\tabcolsep}{2.5pt}
		\centering
		\caption{Comparison of average PSNR and MSE for rotation recovery for 121 simulations.} \label{tab:rot}
		\scriptsize
		\begin{tabular}{ rrrrrrr }
			\toprule
			\multirow{2}{*}{Image} & \multicolumn{3}{c}{Vandewalle \cite{vandewalle2007super}} & \multicolumn{3}{c}{Proposed}  \\
			\cline{2-7} \\[-2ex]
			& PSNR & MSE & Time (s) & PSNR & MSE & Time (s)\\
			\midrule
			\textit{a} & 32.83 & 6.59 & 0.16 & 42.94 & 1.75 & 0.49\\
			\textit{b} & 37.53 & 4.09 & 0.16 & 43.53 & 1.81 & 0.49\\
			\bottomrule
		\end{tabular}
	\end{table}
\end{center}

Table \ref{tab:GT} summarizes some of the results for our translational method with simulated data, where the results are compared with the ground truth (GT) and other baseline methods; i.e. \cite{keren1988image}, \cite{Guizar-Sicairos_2008}, \cite{szeliski1994hierarchical}, in terms of estimated shifts, peak signal-to-noise ratio (PSNR), and mean square error (MSE). 
Since the expressions derived in Section \ref{sec:subpixel} are exact for any arbitrary shift that can be expressed as positive or negative integer powers of 2, in the noise-free case, exact or near-exact solutions can be achieved, which outperforms the state-of-the-art methods. For any other shift amount, we can get arbitrarily close within the closest integer power of 2, which when compared with the state-of-the-art, is still outstanding. 

Table \ref{tab:rot} shows the PSNR, MSE and computational time for our rotation method compared to \cite{vandewalle2007super}, averaged over 121 simulations. Although our technique can recover any rotation angle, since Vandewalle's method \cite{vandewalle2007super} recovers only angles in the range $[-30,30]$, in order to be fair, we compared our results for every $0.5^\circ$ in that range.

We also ran our scale recovery method for 50 images with scale amounts ${1/4, 1/2, 1, 2, 4}$. All experiments returned the exact scale in under $0.09$ seconds. Since wavelet transform downsamples images by $2$ in every level, we can only recover scales that are multiples of $2$. 

Results obtained for combination of transformations can be seen in Tables \ref{tab:rottrans} and \ref{tab:rottransscale}. While Table \ref{tab:rottrans} shows comparisons to Vandewalle's method for rotation and translation, Table \ref{tab:rottransscale} presents our results obtained for several combinations of scale, rotation and translation. These tables also confirm that our method is accurate and outperforms or at least matches state-of-the-art.

\subsection{Optimal Parameters}\label{sec:optpara}
\begin{figure}[h]
	\begin{minipage}{\linewidth}
		\centering
		\centerline{\includegraphics[width=10.4cm]{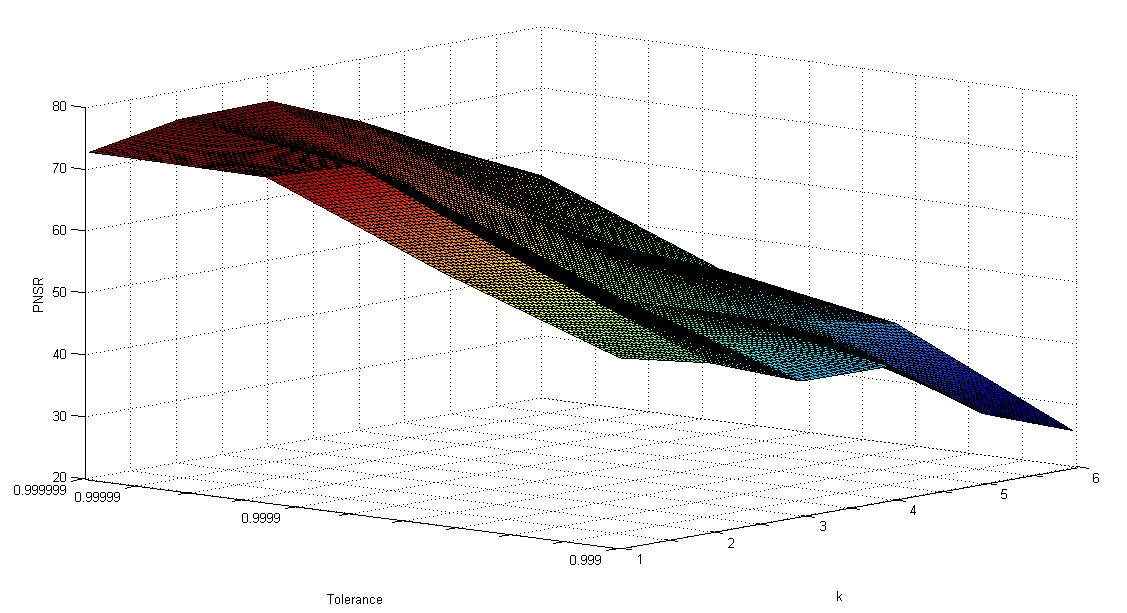}}
	\end{minipage}
	\caption{Comparison of $k$ (x axis) and $\tau$ (y axis) with PSNR (z axis) for average of 50 images for GT shift of $(0.33, -0.33)$.}\label{fig:tau_k_psnr}
\end{figure}

In order to find the appropriate constants $\tau$ and $k$ for translational shift, which are the measures of accuracy (tolerance for cross correlation function) and reduction level of Haar transform, respectively, and show the accuracy of the proposed method, we tested our algorithm with 50 simulated test images for shift amount $(0.33, -0.33)$. Results after removing the outliers (when a local maxima is reached) are shown in Fig. \ref{fig:tau_k_psnr}, where $PSNR = Inf$ is demonstrated as $100$. As seen in the figure, the constants $\tau$ and $k$ can be adapted depending on the trade-off between time complexity and PSNR. 

In case of the most general similarity transformation, $k$ is decided based on the recovered scale $\hat \sigma$ by choosing $k=1$ if $scale < 1$ or $k=\hat \sigma+1$ otherwise.

\begin{center}
	\begin{table*}[!h]\setlength{\tabcolsep}{2.5pt}
		\centering
		\caption{Comparison of PSNR, MSE, and time for rotation and translation recovery.} \label{tab:rottrans}
		\scriptsize
		\begin{tabular}{ rrrrrrrrrr }
			\toprule
			\multirow{2}{*}{Image} & \multirow{2}{*}{Exact $(x,y,\theta)$} & \multicolumn{4}{c}{Vandewalle \cite{vandewalle2007super}} & \multicolumn{4}{c}{Proposed}  \\
			\cline{3-10} \\[-2ex]
			& & Estimate & PSNR & MSE & Time (s) & Estimate & PSNR & MSE & Time (s)\\
			\midrule
			\textit{a} & $(0.5, -0.25, 20)$ & $(0.8, -0.5, 19.8)$ & 23.2 & 16.4 & 0.1 & $(0.5, -0.25, 20)$ & 25.2 & 13.07 & 5.55\\
			\textit{b} & $(-0.375, -0.375, -10)$ & $(-0.337, -0.62, -10.2)$ & 21.17 & 19.97 & 0.09 & $(-0.406, -0.375, -10)$ & 22.05 & 19.7 & 23.4 \\ 
			\textit{c} & $(-0.4375, 0.875, -30)$ & $(-0.64, 0.526, -30)$ & 21.01 & 19.02 & 0.09 & $(-0.39, 0.875, -30.3)$ & 20.2 & 20.86 & 66\\
			\bottomrule
		\end{tabular}
	\end{table*}
\end{center}

\begin{center}
	\begin{table}[!h]\setlength{\tabcolsep}{1.5pt}
		\centering
		\caption{Our results for scale, rotation and translation.} \label{tab:rottransscale}
		\scriptsize
		\begin{tabular}{ rrrr }
			\toprule
			\multirow{2}{*}{Img} & \multirow{2}{*}{Exact $(x,y,\theta,\sigma)$} & \multicolumn{2}{c}{Results}  \\
			\cline{3-4} \\[-2ex]
			& & Estimate & Time (s) \\
			\midrule
			\textit{a} & $(0.5, 0.25, -50, 2)$ & $(0.5, 0.25, -50.1, 2)$ & 25.7 \\
			\textit{a} & $(0.5, 0.25, 50, 2)$ & $(0.5, 0.25, 49.8, 2)$ & 4.94 \\
			\textit{b} & $(-0.25, 0.25, 10, 1/4)$ & $(-0.28, 0.28, 10.2, 1/4)$ & 105.3 \\
			\textit{c} & $(-0.5, -0.375, 30, 1/2)$ & $(-0.5, -0.375, 30.1, 1/2)$ & 93.6 \\
			\bottomrule
		\end{tabular}
	\end{table}
\end{center}

\subsection{Validation on Real Data}\label{sec:realdata}
In order to ensure the accuracy of our method, real world images were also utilized as input. Results for real world examples (\textbf{d}, \textbf{e} and \textbf{f} in Fig. \ref{fig:testim}) including comparisons with the state-of-the-art methods \cite{evangelidis2008parametric} and \cite{vandewalle2007super} are summarized in Table \ref{tab:real}. Since the GT for the used images is not known, the results are compared using PSNR and MSE as it is common practice in the literature. All methods are given the same input, where smaller image regions are used to adopt image sizes to work with our method as described in Section \ref{sec:translation}. As seen in Table \ref{tab:real}, our method outperforms the baseline methods in real world examples in most cases as well.

\subsection{The Effect of Noise and Sparseness}\label{sec:sparse}
\begin{figure}[t]
	\centering
	\centerline{\includegraphics[width=10.8cm]{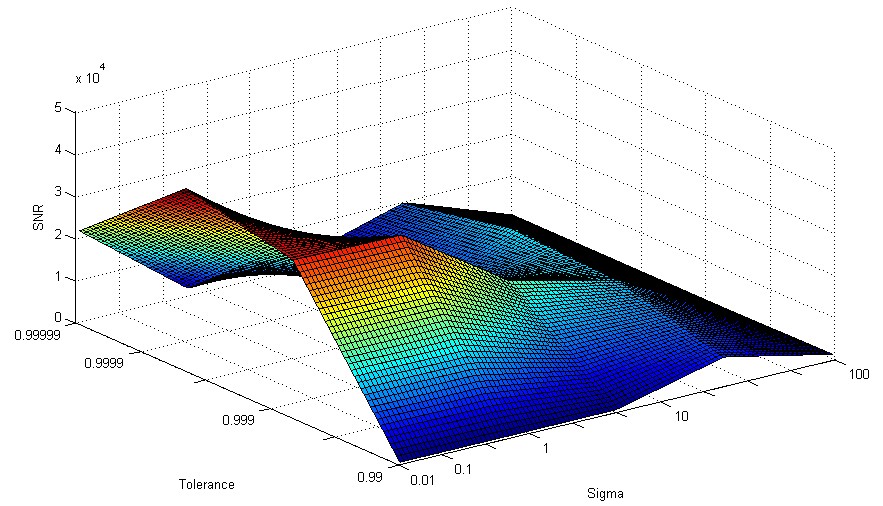}}
	\caption{Average PSNR (z axis) compared with changing $\tau$ (y axis) and $\sigma$ (x axis), for horizontal axis.} \label{fig:noise}
\end{figure}

Our proposed approaches for scale and rotation estimation already suppress noise by hard wavelet thresholding. Therefore, here we discuss only noise in translation estimation. In Table \ref{tab:noise}, a comparison of the proposed method with \cite{Foroosh_etal_2002} and \cite{he2007nonlinear} under noisy conditions is presented. By adapting $\tau$, based on the level of noise and cross validation, very accurate shift values can be achieved. It can be concluded from Table \ref{tab:noise} that our method performs well in suppressing Gaussian noise, which also is superior to the state-of-the-art. In order to show the accuracy under noisy conditions, the proposed algorithm is tested for 50 images with 50 different shift amounts for each image, with Gaussian noise. Results, after removing outliers, are shown in Fig. \ref{fig:noise} for average SNR with respect to $\tau$ and $\sigma$. 

Since our method works entirely in-band (i.e. using only detail coefficients), the method is particularly applicable to wavelet encoded imaging. Moreover, our approach can work with a sparse subset of coefficients, e.g. compressed sensing of wavelet-encoded images. Since our scale and rotation recovery methods already use sparse coefficients (i.e. hard-thresholded wavelet coefficients), we experimented on translational shifts under sparseness. We tested our method as the level of sparseness varied from 2\% to 100\% of detail coefficients, for several simulated images and different shifts. We then fitted a model to the average results to evaluate the trend which is shown in Fig. \ref{fig:sparse}-a. It can be noticed that even at very sparse levels of detail coefficients, the method is stable with an average PSNR above 46dB. Beyond 50\% sampled detail coefficients, the PSNR grows exponentially. 

\begin{center}
	\begin{table*}[h]
		\centering
		\setlength{\tabcolsep}{5pt}
		\caption{Comparison of our method with other methods for real world examples from \cite{CMU} and \cite{UCSC} in PSNR and MSE.} \label{tab:real}
		\scriptsize
		\begin{tabular}{l@{\hspace{5pt}} *{9}{r}}
			\toprule
			\multirow{2}{*}{Dataset}& \multirow{2}{*}{Reference img.}& \multirow{2}{*}{Sensed img.}& \multicolumn{2}{c}{Vandewalle \cite{vandewalle2007super}} & \multicolumn{2}{c}{Evangelidis \cite{evangelidis2008parametric}} & \multicolumn{2}{c}{Proposed} \\
			\cline{4-9}
			& & & PSNR & MSE & PSNR & MSE & PSNR & MSE \\
			\midrule
			Artichoke & 1 & 2 & 26.88 & 11.59 & 31.5 & 6.78 & 31.8 & 6.72  \\
			Artichoke & 27 & 28 &  26.86 & 11.17 & 42.08 & 1.93 & 31.06 & 6.92\\
			CIL & HorizR0 & HorizR1 & 24.02 & 13.2 & 12.4 & 50.4 & 24.73 & 12.66 \\
			CIL & VertR4 & VertR5 & 20.66 & 21.57 & 22.2 & 18.46 & 20.75 & 22.5 \\
			MDSP Bookcase 1 & 2 & 3 & 26.58 & 11.90 & 12.52 & 60.31 & 25.10 & 14.11 \\
			\bottomrule
		\end{tabular}%
	\end{table*}%
\end{center}

\begin{table}[t]
	\begin{center}
		\caption{Comparison of results for noisy environments with "Pentagon" image for $(0.25, 0.75)$ shift.} \label{tab:noise}
		\scriptsize
		{		\begin{tabular}{ llll }
				\toprule
				SNR & Foroosh \cite{Foroosh_etal_2002} & Chen \cite{Chen_Yap2007} & Proposed \\
				\midrule
				10 dB & 0.38 0.65 & 0.29 0.68 & 0.25 0.75\\
				20 dB & 0.31 0.71 & 0.28 0.74 & 0.25 0.75\\
				30 dB & 0.30 0.73 & 0.27 0.74 & 0.25 0.75\\
				40 dB & 0.29 0.74 & 0.27 0.74 & 0.25 0.75\\
				\bottomrule
			\end{tabular} } {}
		\end{center}
	\end{table}

\subsection{Computational Complexity and Convergence Rate}\label{sec:comp}
Time complexity of our method depends on in-band shifting, parameter selection, and the level of sparseness. In-band shifting method in Section \ref{sec:subpixel}, has a complexity of $O((L/2^{N-k+1} )^2)$ for all $k=1...N$, where $L$ is size of the image (or a sparsified version). Parameter selection also affects the complexity since when $\tau$ is higher, the method attempts to match the images with higher accuracy, which would increase the run time. We provide running time of our method with comparisons in Tables \ref{tab:rot}, \ref{tab:rottrans} and \ref{tab:rottransscale} on a machine with 2.7 GHz CPU and 8 GB RAM. 

\begin{figure}[t]
		\begin{minipage}{.45\linewidth}
			\centering
			\centerline{\includegraphics[width=8.2cm]{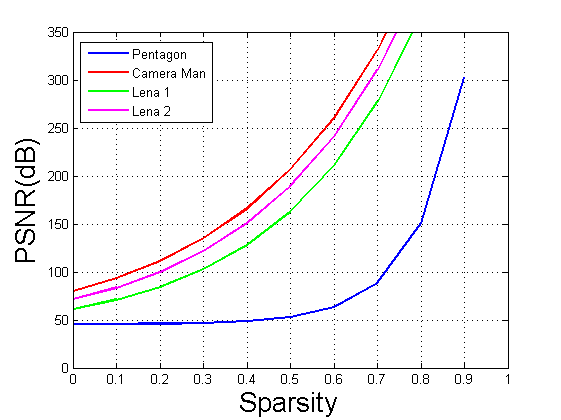}}
			\centerline{\textbf{a}}\medskip
		\end{minipage}
		\hfill		
		\begin{minipage}{.45\linewidth}
			\centering
			\centerline{\includegraphics[width=6.8cm]{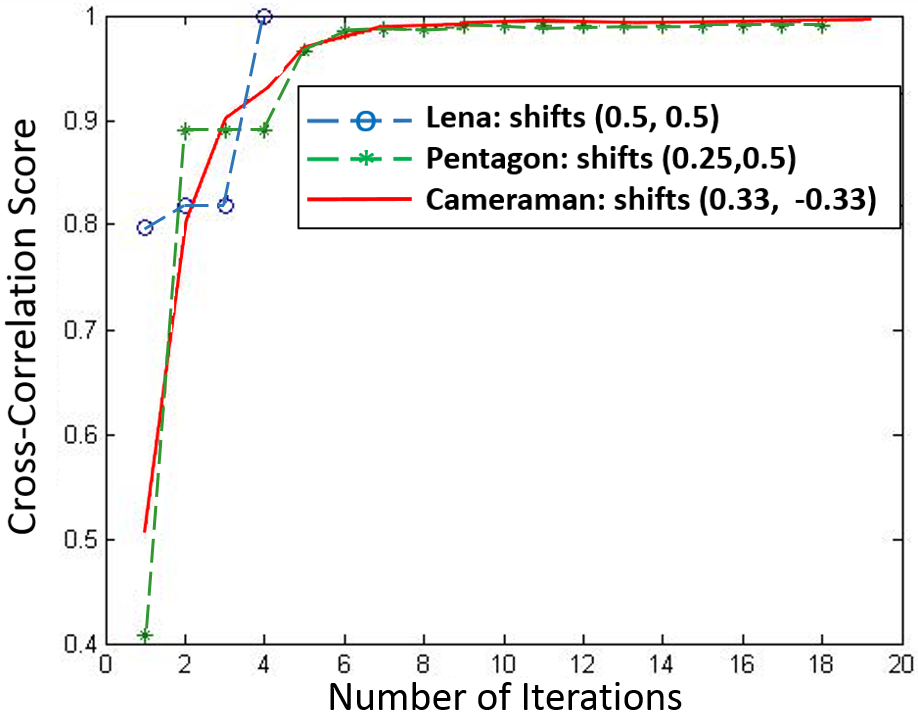}}
			\centerline{\textbf{b}}\medskip
		\end{minipage}
		\caption{\textbf{a} Average PSNR as a function of percentage of detail coefficients (level of sparsity) used to register for Pentagon, Cameraman, and two different shifts of Lena. In all cases, the worst registration PSNR when using only 2\%-7\% of detail coefficients was above 46dB \textbf{b} Examples illustrating the convergence to optimal cross-correlation as a function.}\label{fig:sparse}
	\end{figure}

Fig. \ref{fig:sparse}-b demonstrates the convergence of our method to the global cross-correlation maximum for Lena image with GT $(0.5, 0.5)$ in blue circles, Pentagon image with GT $(0.25, 0.5)$ in green stars, and Cameraman image with GT $(0.33, -0.33)$ in red line. The convergence is visibly exponential and therefore we get a very rapid convergence to the solution. 

\section{Conclusion} \label{sec:conclusion}
A sub-pixel registration technique for sparse Haar encoded images is proposed. Only a sparse set of detail coefficients are sufficient to establish the cross-correlation between images for scale, rotation, and translation recovery. Our registration process is thus performed solely in-band, making the method capable of handling both in-band registration for wavelet-encoded imaging systems, and sparsely sensed data for a wavelet-based compressive sensing imager. Moreover, our method conveniently decouples scale, rotation and translation parameters, while exploiting Haar wavelet's important features, such as multi-resolution representation and signal energy localization. Our method does not use image interpolation for estimating the registration parameters, since the exact set of in-band equations are derived for establishing the registration and fitting the parameters. Although the run time of our method is higher than compared methods, we achieve far better accuracy as a reasonable trade-off. Overall, our results show superior performance, and outperform the baseline methods in terms of accuracy and resilience to noise. 


{
\bibliographystyle{ieeetr}
\bibliography{foroosh,corresreg,IEEEabrv}
}

\end{document}